\documentclass[twocolumn]{autart}
\usepackage[utf8]{inputenc}
\usepackage[T1]{fontenc}
\usepackage{latexsym, amsmath, amssymb, amsfonts,
            upgreek, mathrsfs, mathtools}
\usepackage{arydshln}   
\usepackage{empheq}     
\usepackage{enumitem}   

\usepackage{algorithm}
\usepackage{algpseudocode}

\usepackage{float, graphicx, caption, subcaption}
\usepackage{wrapfig}
\usepackage{rotating}
\usepackage{multirow,booktabs}
\usepackage[usenames,dvipsnames]{xcolor}
\usepackage{tikz}
\usepackage{comment}

\newcommand{\mvar}[1]{\operatorname{MVAR}(#1)}

\newcommand{\argmin}{\operatorname{arg\,min}}

\begin{document}
\begin{frontmatter}

\title{A system identification approach to clustering vector
  autoregressive time series}

\thanks[footnoteinfo]{This paper was not presented at any IFAC meeting.}

\author[hust,unsw]{Zuogong Yue}\ead{z\_yue@hust.edu.cn},
\author[hust]{Xinyi Wang}\ead{m202173156@hust.edu.cn},
\author[unsw]{Victor Solo}\ead{v.solo@unsw.edu.au}

\address[hust]{School of Artificial Intelligence and Automation,
    Huazhong University of Science and Technology,
    China}
\address[unsw]{School of Electrical Engineering and Telecommunications,
  UNSW Sydney,
  NSW, Australia}


\begin{abstract}
  Clustering of time series based on their underlying dynamics is keeping
  attracting researchers due to its impacts on assisting complex system
  modelling.
  Most current time series clustering methods handle only scalar time
  series, treat them as white noise, or rely on domain knowledge for
  high-quality feature construction, where the autocorrelation
  pattern/feature is mostly ignored.
  Instead of relying on heuristic feature/metric construction, the system
  identification approach allows treating vector time series clustering by
  explicitly considering their underlying autoregressive dynamics.
  We first derive a clustering algorithm based on a mixture autoregressive
  model. Unfortunately it turns out to have significant computational
  problems. We then derive a `small-noise' limiting version of the
  algorithm, which we call k-LMVAR (Limiting Mixture Vector
  AutoRegression), that is computationally manageable. We develop an
  associated BIC criterion for choosing the number of clusters and model
  order. The algorithm performs very well in comparative simulations and
  also scales well computationally.
\end{abstract}

\end{frontmatter}

\section{Introduction}
\label{sec:introduction}


Cluster analysis, or clustering, is a common technique for exploratory
data mining and statistical data analysis, whose study has a long
history dating back to 1932 \cite{Driver1932}.
%
%
Nowadays many applications necessitate clustering autocorrelated time
series, e.g., stock prices, locations of mobile robots, speech signals,
ECG, etc.
%
%
The introduction first reviews popular algorithms in machine learning
for time series clustering, which rely on either prescribed similarity
metrics or observable features from time series.
{\color{black} Applications in systems and control motivate the clustering
  of time series as the essential preprocessing for piecewise linear models
  to represent complex dynamics \cite{Ferrari-Trecate2003} or for
  clustering-based control system analysis and design
  \cite{Niazi2023,Mattioni2022}. These motivations drive us to cluster time
  series by their underlying governing dynamics rather than man-made
  similarity metrics. This perspective then leads to an interplay of
  clustering and cluster-specific system identification, as discussed in
  this paper.}

\subsection{Time-series clustering in machine learning}
Although the collection of `dynamic' data i.e. time-series (e.g.
wearable devices) has grown rapidly in recent decades, the associated
machine learning clustering literature is still relatively thin
especially where vector time series are concerned. The available methods
on clustering temporal data can roughly be categorised into two types:
distance-based methods and feature-based methods. As indicated above
these methods generally do not explicitly incorporate the
autocorrelation feature.

The distance-based methods rely on careful design of similarity metrics
that manage to reveal time-series characteristics. Perhaps the most
widely used measure is \emph{dynamic time warping} (DTW)
\cite{Sakoe1978}. A global averaging method has been developed for DTW,
called \emph{DTW barycenter averaging} (DBA), which allows an easy
integration with k-Means for time series clustering
\cite{Petitjean2011}. The \emph{K-Spectral Centroid} (K-SC) algorithm
was developed in \cite{Yang2011} to effectively finds cluster centroids
with a new measure that is invariant to scaling and shifting. More
recently \cite{Paparrizos2015,Paparrizos2017} presented a novel method
called \emph{k-Shape} by considering `shapes' of time series to
construct a metric for dissimilarity \cite{Ives2016}.

The feature-based methods refer to such algorithms that cluster the
extracted `feature' representations of time series. The approach has two
parts: feature learning and classical clustering; while, in most
methods, these two stages are jointly and iteratively optimized. The
traditional techniques, like \cite{Guo2008,Li2019c}, use ICA or PCA to
extract low-dimensional features for clustering. Another family of
clustering methods builds upon a widely used feature called
\emph{shapelet}, such as \emph{u-Shapelets}
\cite{Zakaria2012,Ulanova2015,SiyouFotso2020} and \emph{salient
  subsequence learning} \cite{Zhang2018b}, which are based on the rapid
development of fast, robust and scalable algorithms for \emph{shapelet
  learning}. The recent popular approaches integrate into clustering
anonymous features extracted by heavy black-box modelling, while feature
interpretability mostly becomes an issue. For example, \emph{deep
  temporal clustering} (DTC) uses an autoencoder and a clustering layer
to join feature learning and clustering \cite{Madiraju2018}, and an
improved DTC with a fake-sample generation strategy, named DTCR
\cite{Ma2019a}.

\subsection{Clustering in a perspective of system dynamics }
\label{sec:persp-syst-dynam}


In physical applications feature-based clustering may not be
sufficient. One needs a physical interpretation of the classified
classes. This is where autoregressive type models are advantageous since
they offer interprations in terms of time-constants, resonant
frequencies, feedback etc. For instance, in the study
\cite{Bulteel2016}, psychologists model time-series data on
depression-related symptoms, using the fitted dynamics to identify and
understand different types of psychiatric disorders.

The (vector) \emph{autoregressive} (AR/VAR) model is a classical
technique to describe univariate/multivariate time series, which is a
reliable model widely used in engineering, economics, biology and medial
applications. To model multimodality in marginal or conditional
distributions of scalar time series, the \emph{mixture autoregressive}
(MAR) model was developed by \cite{Wong2000}. It was further extended by
\cite{Wong2001,Wong2001a,Lanne2003}. Applying autoregressive techniques
is not a new idea to provide an explainable clustering method for time
series data, e.g., \cite{Corduas2008} and \cite{Xiong2002,Xiong2004} are
for univariate time series. Also \cite{Yue2020} recently developed a new
algorithm based on \cite{Xiong2004} for univariate time series, which
scales well for large-scale clustering tasks.

To deal with multivariate time series, there have been attempts at
developing methods to cluster VAR models in application literature. A
hard-clustering algorithm was proposed in \cite{Bulteel2016}, which,
however, only focuses on a special case, the first-order VAR model
$\mathrm{VAR}(1)$. Further \cite{Takano2020} did a comparative study on
the algorithm of \cite{Bulteel2016} and a straightforward two-step
approach (first estimating VAR model and then clustering via Gaussian
mixture models) on limited features. Similarly for the specialised
$\mathrm{VAR}(1)$ model, the work of \cite{Huang2012} provided a full
Bayesian sparse method to solve bi-clustering problem. The direct
derivation of hard-clustering of general VAR model is particularly
challenging; that is why we chose to resort to soft clustering as a
starting point.

A corresponding mixture model for multivariate time series is needed in
order to extend the univariate case (e.g., \cite{Xiong2002,Xiong2004}).
We use the mixture vector autoregressive (MVAR) model from
\cite{Fong2007}, and its periodic version from \cite{Bentarzi2014}. This
paper may be seen as a multivariate extension of our previous work
\cite{Yue2020}, although everything has to be redone from the ground up
and significant new problems have to be overcome.

%
Instead of the regular path from hard clustering to soft clustering, our
study goes in the opposite direction.
First, using an existing vector time series mixture model we develop an
EM based soft-clustering algorithm, \emph{cMVAR}. The computation turns out to be
prohibitive and to be also plagued by numerical underflow problems. So
we take a small-noise limit and end up with a hard-clustering method,
\emph{k-LMVAR}, which is computationally manageable and
does not have numerical issues. We have not yet found a general
derivation that dispenses with the EM setting.

\subsection{Contributions}
{\color{black}
We develop a new scalable clustering algorithm k-LMVAR for
autocorrelated vector time series. In particular we make explicit use of
the fundamental feature of (vector) time series that has so far been
largely ignored, autocorrelation.
We first develop a new EM type clustering algorithm cMVAR by
applying existing mixture VAR model. But it turns out to have major
computational issues, which consequently limit the extent of its
applicability. By taking a small noise limit we produce a
simplified algorithm k-LMVAR that overcomes these limits. Its
convergence can be seen by recognising its coordinate
descent procedure,
and a BIC criterion is proposed to choose the number of clusters and VAR orders.
%
Its performance has been shown in comparative studies on synthetic data,
together with its superiority by incorporating autocorrelation features. }


\section{Problem Description}
\label{sec:prob-descrip}


Consider a set of $N$ time series
$\mathbb{Y} \triangleq \{{Y}^{(1)}, \dots, {Y}^{(N)}\}$, where each
${Y}^{(n)}$ ($1\leq n \leq N$) denotes a time series of $m$-dimensional
vectors $Y_1^{(n)}, \dots, Y_T^{(n)}$. The time series clustering
problem is to group these $N$ number of time series into $K$ clusters
$\mathbb{C} = \{C_1, \dots, C_K\}$ ($K \leq N$).
%
The classical treatment is to define measures that quantify dissimilarity
between time series, or to extract non-temporal features from time series
and then cast it into a standard clustering problem. Nevertheless, the
system perspective demands to clustering time series with respect to their
underlying dynamics.

Our system perspective considers a set of $N$ time series each of which is
generated by an LTI system. The task is to cluster them into $K$ groups
according to their underlying dynamics similarity. To be specific,
supposing $\{H_n(s)\}_{n=1}^N$ denotes the systems that correspond to $N$
time series $\{\mathcal{D}_n\}_{n=1}^N$, we classify $\{H_n(s)\}_{n=1}^N$
is into $K$ groups according to similarity between systems. This refers to
such a methodology that the essential features used for time-series
clustering is interpretable in a view of dynamics point.

A naive idea is to solve this problem by two stages: 1) identifying $N$
models $\{\hat{H}_n(s)\}_{n=1}^N$ from each time series; and 2) clustering
them by comparing distances to $K$ `mean' models
$\bar{H}_1, \dots, \bar{H}_K$. However, this may not be the best choice due
to the errors accumulated at stage 1 that considerably affect clustering
accuracy, which has been shown in comparative studies (see
Section~\ref{sec:experiments}).
Enlighten by classic clustering algorithms, it can be improved by combining
two-stage operations, i.e. iterating between cluster-specific model
estimation and cluster label updating. The correctly labelled times series
can be used to refine their cluster-specific model estimation; while, on
the other hand, better models help to precisely classify time series. The
resultant performance then is improved over iterations. Unfortunately, such
an iterative algorithm is not easy to develop due to challenges on
construction of dissimilarity measure for time series. The later sections
will firstly propose an algorithm that realises this idea but fails to
scale, from which a new scalable version will be derived.


\section{Initial Clustering Algorithm cMVAR}
\label{sec:ts-clustering}

To realise our idea on algorithm development, we need to find an
appropriate model that could describe mixture of time series. A model,
mixture vector autoregressive (MVAR) in \cite{Fong2007}, is introduced,
although it was originally created to model a single time series that
exhibits multimodality either in the marginal or the conditional
distribution. Then based on MVAR model we propose our initial clustering
algorithm {cMVAR}, which does model updating and clustering in an
iterative manner.

\subsection{Preliminaries: MVAR model}
\label{sec:preliminaries}

An MVAR model consists of a mixture of $K$ Gaussian VAR components
for an $m$-dimensional vector $Y_t$, denoted by
$\mvar{m,K; p_1, p_2, \dots, p_K}$, is defined as
\begin{equation*}
  \begin{array}{r@{\;}l}
    F(Y_t \mid \mathcal{F}_{t-1}) =
    &\sum_{k=1}^K \alpha_k \Phi \Big(\Omega_k^{-1/2} (Y_t - \Theta_{k0}
      - \Theta_{k1}Y_{t-1} \\
    &-\Theta_{k2}Y_{t-2} - \cdots - \Theta_{kp_k}Y_{t-p_k}) \Big),
  \end{array}
\end{equation*}
where $p_k$ is the VAR order for the $k$-th component, $\mathcal{F}_{t-1}$
indicates the information given up to time $t-1$, $\Phi(\cdot)$ is the
multivariate cumulative distribution function of the Gaussian distribution
with mean zero and identity covariance matrix, $\alpha_k$ is the
probability for the $k$-th component, $\Theta_{k0}$ is an $m$-dimensional
vector, $\Theta_{k1},\dots,\Theta_{kp_k}$ are $m \times m$ coefficient
matrices for the $k$-th component, and $\Omega_k$ is the $m \times m$
covariance matrix for the $k$-th component, $k=1,\dots,K$. To avoid
noidentifiability due to the interchange of component labels, it is assumed
that $\alpha_1 \geq \alpha_2 \geq \cdots \geq \alpha_K \geq 0$, and
$\sum_{k=1}^K \alpha_k = 1$.
To estimate model parameters from a multimodal time series, the EM
algorithm can be applied to solve the maximum likelihood estimation (see
\cite{Fong2007}). Note that the MVAR model was originally inferred from
a single multi-modal time series, which is not for clustering.

{\color{black}

  The development of \emph{cMVAR} for clustering is simply the
  identification of MVAR model using the set of $N$ time series
  $\mathbb{Y}$. The values of $m, K$ and model orders $p_1, \dots, p_K$ are
  known \emph{a priori} or be chosen by users. The parameters pending to be
  estimated consist of mixture coefficients $\alpha_{k}$ and and
  component-specific model parameters
  $\left\{ \Theta_{k0}, \dots, \Theta_{kp_k} \right\}$ for all $k$. The
  clustering labels are depicted by the latent variables in the EM-based
  estimation algorithm, given later in Section~\ref{subsec:mvar-tsclust}.

}

\subsection{cMVAR  clustering algorithm}
\label{subsec:mvar-tsclust}


This section develops this EM based clustering algorithm, \emph{cMVAR}.
Suppose that each time series ${Y}^{(n)}$ in $\mathbb{Y}$ is generated from
one component (that is, a VAR model) of the MVAR. Let
$p \triangleq \max \{p_1, \dots, p_K\}$, and, suppose that the data points
earlier than $t = p$ have been given. The log likelihood for the whole data
$\mathbb{Y}$ is given by
\begin{equation}
  \label{eq:lld}
  \hspace*{-2mm}
  \log L = \textstyle\sum_{n=1}^N \sum_{t=p+1}^T \log
  \left( \sum_{k=1}^K \alpha_k f_{k}(\mathbf{e}_{nkt}, \Omega_k) \right)
\end{equation}
where
\begin{align*}
  f_k(\mathbf{e}_{nkt},\Omega_k) &= \textstyle (2\pi)^{-\frac{m}{2}} |\Omega_k|^{-\frac{1}{2}}
      \exp\left( -\frac{1}{2} \mathbf{e}_{nkt}^T \Omega_k^{-1}
                                   \mathbf{e}_{nkt}\right), \\
  \mathbf{e}_{nkt} & \begin{array}[t]{r@{}l}
                       &= Y_t^{(n)} - \Theta_{k0} - \Theta_{k1}Y_{t-1}^{(n)}
                         -\cdots- \Theta_{kp_k}Y_{t-p_k}^{(n)} \\[4pt]
                       &\triangleq Y_t^{(n)} - \tilde{\Theta}_k X_{kt}^{(n)},
                     \end{array}\\
  \tilde{\Theta}_k &= \left[\Theta_{k0}, \Theta_{k1}, \dots, \Theta_{kp_k} \right],\\
  X_{kt}^{(n)} &= \left[ {1}, Y_{t-1}^{(n)T}, Y_{t-2}^{(n)T}, \dots,
                 Y_{t-p_k}^{(n)T} \right]^T
\end{align*}
for $k=1,\dots,K$. For convenience, we use $\alpha$ to denote the
collection of $\alpha_1, \dots, \alpha_K$; $\tilde{\Theta}$ to denote the
collection of $\tilde{\Theta}_1, \dots, \tilde{\Theta}_K$; and $\Omega$ to
denote the collection of $\Omega_1, \dots, \Omega_K$.

To derive the EM algorithm, let the latent variable be defined by
$Z_n = (Z_{n,1},\dots,Z_{n,K})^T$ for $n=1,\dots,N$, where $Z_{n,k} = 1$
($1 \leq k \leq K$) if the time series ${Y}^{(n)}$ comes from the $k$-th
component; otherwise, $Z_{nk} = 0$. The log likelihood for the complete
data $({Y}^{(1)}, Z_1, \dots, {Y}^{(N)}, Z_N)$ is, omitting the constant
terms, given by
\begin{equation}
  \label{eq:lld-compl-data}
  \begin{array}{r@{\;}l}
    \log\mathcal{L} =
    &\sum_{n=1}^N \sum_{k=1}^K Z_{n,k}
    \Big(
      (T-p)\log(\alpha_k)\\[4pt]
    &-\frac{T-p}{2}\log|\Omega_k| - \frac{1}{2} \sum_{t=p+1}^T \mathbf{e}_{nkt}^T \Omega_k^{-1} \mathbf{e}_{nkt}
    \Big).
  \end{array}
\end{equation}
To estimate the parameters by maximising the log likelihood function, the
iterative EM procedure is proposed, summarised as follows, for
$n=1,\dots,N$,
\begin{itemize}
\item E-step:
  \begin{align}
    \label{eq:E-tau}
    \tau_{n,k} &= \frac{\alpha_k |\Omega_k|^{-\frac{T-p}{2}}
      \exp\left(-\frac{1}{2} \psi_{n,k} \right)
    }{ \sum_{k=1}^K \alpha_k |\Omega_k|^{-\frac{T-p}{2}}
      \exp\left(-\frac{1}{2} \psi_{n,k} \right) }, \\
    \label{eq:E-psi}
    \psi_{n,k} &= \textstyle \sum_{t=p+1}^T \mathbf{e}_{nkt}^T
                 \Omega_k^{-1}  \mathbf{e}_{nkt};
  \end{align}

\item M-step:
  \begin{align}
    \label{eq:M-alpha}
    \alpha_k &= \frac{1}{N} \textstyle\sum_{n=1}^N \tau_{n,k},\\
    \label{eq:M-theta}
    \tilde{\Theta}_k^T &=
                         \begin{array}[t]{@{}l@{}l}
                           &\left[ \sum_{n=1}^N \tau_{n,k} \big( \sum_{t=p+1}^T X_{kt}^{(n)}
                             X_{kt}^{(n)T}  \big) \right]^{-1} \\[4pt]
                           &\left[ \sum_{n=1}^N \tau_{n,k} \big( \sum_{t=p+1}^T
                             X_{kt}^{(n)}Y_{t}^{(n)T} \big) \right],
                         \end{array} \\
    \label{eq:M-omega}
    \Omega_k &=
               \begin{array}[t]{@{}l@{}l}
                 &\left[ (T-p)\sum_{n=1}^N \tau_{n,k} \right]^{-1} \\[4pt]
                 &\left[ \sum_{n=1}^N \tau_{n,k}
                   \big( \sum_{t=p+1}^T \mathbf{e}_{nkt} \mathbf{e}_{nkt}^T \big)
                   \right].
               \end{array}
  \end{align}
\end{itemize}
The clustering labels are determined by examining the convergence values
of $\tau_{n,k}$, which is the posterior probability that component $k$
was responsible for generating time series ${Y}^{(n)}$ and, in other
words, is the probability that ${Y}^{(n)}$ belongs to the $k$-th
cluster.

%
However, for scalar time series, we found that clustering with mixture
AR/ARMA models has major computational problems \cite{Yue2020}. The
algorithms do not scale up and they also suffer from numerical underflow
issues \cite{Yue2020}.
{\color{black} All this is exacerbated in the vector case. The cMVAR
  algorithm can easily fail to function when $m$ or $T$ gets large. The
  underflow issue happens in the computation of $\tau_{n,k}$ in
  \eqref{eq:E-tau}, where the summation over time $t$ and variable
  dimension $m$ leads to large $\psi_{n,k}$ and then could result in
  particularly small values after the exponential operator. Although we
  could apply some tricks, e.g. normalising every term in the denominator
  by the numerator, the underflow issue is still inevitable for
  `high-quality' time series that is of large time length $T$. To resolve
  these intrinsic drawbacks, the key is to remove the formula of
  \eqref{eq:E-tau} from the iteration that is the exact cause of numerical
  underflow issues. We achieve this simplification by taking a small-noise
  limit as now described. This leads to a hard-clustering method from
  cMVAR. }

\section{Scalable Algorithm k-LMVAR}
\label{sec:kARs-theory}

In this section we take a small noise `limit' of the cMVAR algorithm to get
a simpler and scalable clustering algorithm. %
{\color{black} The parameters to be estimated are similar to \emph{cMVAR},
  except that mixture coefficients $\alpha_k$ have been removed and
  clustering labels are given directly rather than being examined by
  posterior probability $\tau_{n,k}$.}

Let $\Omega_k$ be written as
$\Omega_k = \gamma_k \tilde{\Omega}_k \coloneqq \gamma_k (\Omega_k /
\det(\Omega_k))$ for $k=1,\dots,K$. We would like to take the limits
when all $\gamma_k$ go to zero at the same speed. To do this in a
coherent way, denote the ratios
$\lambda_k \triangleq {\gamma_k}/{\gamma}$, in which
$\gamma = \max_k \gamma_k$. As $\gamma \rightarrow 0$ with the ratios
$\lambda_k$ being constant in the E-step, we obtain
\begin{equation}
  \label{eq:M-tau-limit}
  \tau_{n,k} \rightarrow \tau_{n,k}^* =
  \begin{cases}
    1 & \text{if } k = \argmin_{\hat{k}} \psi_{n,\hat{k}} \\
    0 & \text{otherwise,}
  \end{cases}
\end{equation}
where $\psi_{n,\hat{k}}$ is given in \eqref{eq:E-psi}.
Now $\tau_{n,k}$ turns to be a binary indicator variable
$\tau_{n,k}^*$. It helps to decrease computation cost by computing over
index sets of each cluster $k$, which is defined by
\begin{equation}
  \label{eq:index-set-cluster-k}
  I_k = \left\{ n: \tau^*_{n,k} = 1; n=1,\dots,N \right\},
\end{equation}
for $k=1,\dots,K$. With all $\tau_{nk}$ in the M-step replaced by $\tau_{nk}^*$,
the method called \emph{k-LMVAR} is given as:
\begin{itemize}
\item label update:
  \begin{align}
    \label{eq:E-kAR-tau}
    \tau_{n,k}^* &=
    \begin{cases}
      1 & \text{if } k = \argmin_{\hat{k}} \psi_{n,\hat{k}}, \\
      0 & \text{otherwise,}
    \end{cases} \\
    \label{eq:E-kAR-psi}
    \psi_{n,\hat{k}} &= \textstyle \sum_{t=p+1}^T \mathbf{e}_{n\hat{k}t}^T
                       \tilde{\Omega}_{\hat{k}}^{-1}  \mathbf{e}_{n\hat{k}t};
  \end{align}

\item parameter update:
  \begin{align}
    \label{eq:M-kAR-theta}
    \tilde{\Theta}_k^T &=
               \begin{array}[t]{@{}l@{}l}
                 &\left[ \sum_{n \in I_k} \big( \sum_{t=p+1}^T X_{kt}^{(n)}
                   X_{kt}^{(n)T}  \big) \right]^{-1} \\[4pt]
                 &\left[ \sum_{n \in I_k} \big( \sum_{t=p+1}^T
                   X_{kt}^{(n)}Y_{t}^{(n)T} \big) \right],
               \end{array}\\
    \label{eq:M-kAR-omega}
    \tilde{\Omega}_k &= \Omega_k / \det(\Omega_k), \\
    \Omega_k &= \frac{1}{(T-p) |I_k|}
               \textstyle
               \left[ \sum_{n \in I_k}
               \big( \sum_{t=p+1}^T \mathbf{e}_{nkt} \mathbf{e}_{nkt}^T \big)
               \right],
  \end{align}
  where $|I_k|$ denotes the number of elements of $I_k$ (i.e., the
  cardinality of set $I_k$), and $p \triangleq \max \{p_1, \dots, p_K\}$.
\end{itemize}
The k-LMVAR might be considered as a hard-assignment version of the
cMVAR algorithm with general covariance matrices. The relation could be
understood by analogy with the Gaussian mixture model versus the
\emph{elliptical k-Means} algorithm in \cite{Sung1998}.

{\color{black} The k-LMVAR algorithm, improves on the cMVAR algorithm in two
  ways. The k-LMVAR removes the computation of \eqref{eq:E-kAR-tau} in
  cMVAR, and hence avoids the underflow issues due to computing exponents
  of very large negative values when the dimension $m$ or time length $T$
  is particularly large. This is the major motivation that leads us to
  k-LMVAR. As a byproduct, regarding the intensive matrix computations in
  \eqref{eq:M-kAR-theta} and \eqref{eq:M-kAR-omega}, in comparison with
  cMVAR, each iterate in k-LMVAR only calls for $\sum_k |I_k| = N$ number
  of computation of \eqref{eq:heavy-matrix-computations}, while each
  iterate in cMVAR demands $NK$ number of computations. This cuts
  computational cost a lot. However, we have to admit that, in simulation,
  we observed that k-LMVAR may require more iterations than cMVAR, for
  small-sized problems that cMVAR can apply.
}

The algorithmic nature of k-LMVAR has been considerably different from
cMVAR, where cMVAR is an EM procedure while k-LMVAR turns a coordinate
descent procedure. We can argue easily the basic convergence of k-LMVAR
by constructing a suitable objective function,
\begin{math}
  f(\tau, \tilde{\Theta}, \tilde{\Omega}) =
  \sum_{n=1}^N \sum_{k=1}^K \tau_{n,k} D_{n,k}(\tilde{\Theta}_k,
  \tilde{\Omega}_k)
\end{math} with an equivalent dissimilarity measure
\begin{math}
  D_{n,k}(\tilde{\Theta}_k, \tilde{\Omega}_k) = (T-p)\log |\tilde{\Omega}_k| +
  \textstyle\sum_{t=p+1}^T \mathbf{e}_{nkt}^T \tilde{\Omega}_{k}^{-1}  \mathbf{e}_{nkt}
\end{math}. %
We then see the k-LMVAR algorithm runs as a coordinate (aka. cyclic)
descent algorithm \cite{Wright2015,Luenberger2008}. It is well known of the
uniqueness in probability 1 of the least squares estimate of VAR and its
noise variance from a time series, if the data is generated by a stable VAR
with full-rank noise variance. Then similar to the classic convergence
analysis for k-Means-type algorithms \cite{Selim1984}, with the singleton
set of VAR estimates for each time series, it can be addressed that the
sequence of objective values is non-increasing and the k-LMVAR algorithm
will stop in a finite number of iterations and guarantee local optimality.

\section{k-LMVAR Computation}
\label{sec:kARs-computation}

This section discusses the computation of k-LMVAR, including fast
computation details; and the complete algorithm is present in
Algorithm~\ref{alg:k-ARs}.

\subsection{Initialisation}
\label{subsec:initialisation}

{\color{black}

  To start the k-LMVAR algorithm
  \eqref{eq:E-kAR-tau}-\eqref{eq:M-kAR-omega}, it demands three parameters
  to be initialisation: clustering labels $\tau$, and cluster-specific
  model parameters $\tilde{\Theta}_k$ and $\Omega_k$, for $k=1,\dots,K$. A
  natural way to initialise $\tau$ is, for each $n=1,\dots,N$, randomly
  choosing one $k$ with $\tau_{n,k} = 1$ and setting all other
  $\tau_{n,\hat{k}} = 0$ ($\hat{k} \neq k$). In terms of VAR-like
  parameters $(\tilde{\Theta}_k, \tilde{\Omega}_k)$ the key is to select
  $K$ pairs of parameters such that they are essentially different. An
  alternative way is to perform standard VAR model estimation of every time
  series $Y^{(n)}$, yielding $(\tilde{\Theta}_n^*, \Omega_n^*)$
  ($n=1,\dots,N$). The onus is to choose $K$ values among $N$ results for
  initialisation. One perspective is to treat it as a standard clustering
  problem, that is, clustering $N$ number of $\tilde{\Theta}_n^*$ (or
  $\Omega_n^*$, or both) into $K$ groups using the classic algorithms as
  k-Means, k-Means++ or Gaussian mixture model. Then arbitrarily choosing
  one $\tilde{\Theta}_n^*$ (and $\Omega_n^*$) from each cluster to
  initialise the cluster-specific parameters for the k-LMVAR algorithm. In
  essence, this is a naive way to perform time-series clustering, which is
  called the \emph{naive 2-step} method in our experiments.

  Due to the nature of convergence that is similar to k-Means-type
  algorithms, the initialisation does have noticeable impacts on clustering
  performance due to no global optimality guarantee. Nevertheless, the
  initialisation of clustering labels is not as significant as k-Menas-type
  methods, as learnt from numerical experiments. On a contrary, whether we
  could find $K$ different initial VARs (not necessarily being accurate)
  usually play an important role, where we have greater flexibility to
  optimise its performance.}

\subsection{Stopping criteria}
\label{subsec:stopping-criterion}



A straightforward choice of stopping criterion is to check whether all
parameters $\tau, \tilde{\Theta}, \tilde{\Omega}$ converge. Instead of
$\tau$, we may consider using $\alpha_k = |I_k| / N$ ($k=1,\dots,K$),
which are real-valued and easy to be checked. Alternatively, a better
criterion is given by the convergence argument. We can instead check
whether the cost $f(\tau, \tilde{\Theta}, \tilde{\Omega})$ converges,
defined in Section~\ref{sec:kARs-theory}.

\subsection{Fast matrix computation}
\label{subsec:fast-computation}

The computation bottleneck of the k-LMVAR algorithm is
\eqref{eq:M-kAR-theta} and \eqref{eq:M-kAR-omega}. To rewrite in
compact forms, let
\begin{equation}
  \label{eq:heavy-matrix-computations}
  \begin{array}{r@{\;}l}
  \mathbf{X}_{nk} &\triangleq
                    \begin{bmatrix}
                      X_{k,p+1}^{(n)} & \cdots & X_{k,T}^{(n)}
                    \end{bmatrix}^T, \\
  \mathbf{Y}_{n} &\triangleq
                   \begin{bmatrix}
                     Y_{p+1}^{(n)} & \cdots & Y_{T}^{(n)}
                   \end{bmatrix}^T,\\
  \mathbf{E}_{nk} &\triangleq
                    \begin{bmatrix}
                      \mathbf{e}_{nk,p+1} & \cdots & \mathbf{e}_{nkT}
                    \end{bmatrix}^T,\\[4pt]
    \mathbf{X}_{nk}^T \mathbf{X}_{nk}
                  &=\sum_{t=p+1}^T X_{kt}^{(n)} X_{kt}^{(n)T}, \\[6pt]
    \mathbf{X}_{nk}^T \mathbf{Y}_{n}
                  &=\sum_{t=p+1}^T X_{kt}^{(n)}Y_{t}^{(n)T}, \\[6pt]
    \mathbf{E}_{nk}^T \mathbf{E}_{nk}
                  &=\sum_{t=p+1}^T \mathbf{e}_{nkt} \mathbf{e}_{nkt}^T,
  \end{array}
\end{equation}
where $\mathbf{X}_{nk}$ is of dimension $(T-p) \times (1+ mp_k)$,
$\mathbf{Y}_{nk}$ of dimension $(T-p) \times m$, and $\mathbf{E}_{nk}$
of dimension $(T-p) \times m$. We thus have
$\mathbf{E}_{nk} = \mathbf{Y}_{nk} - \mathbf{X}_{nk}
\tilde{\Theta}^T_{k}$. Note that, the subscription $k$ in
$\mathbf{X}_{nk}$ is due to the selection of different values for
$p_1, \dots, p_K$. If all $p_k$ is equal to the same value,
$\mathbf{X}_{nk}$ will be irrelevant to $k$, and only the $N$ number of
$\mathbf{X}_{nk}$ need to be computed and cached.

Consider the QR decomposition 
\begin{math}
  \mathbf{X}_{nk} = Q_{nk} R_{nk}
\end{math}
($n = 1,\dots,N; k = 1,\dots,K$),
where $Q_{nk}$ is of dimension $(T-p) \times (1+mp_k)$ and
$Q_{nk}^T Q_{nk} = I$, and $R_{nk}$ is an upper-triangular matrix of dimension
$(1+mp_k) \times (1+mp_k)$. And
$\mathbf{Y}_{Q_{nk}} \triangleq Q_{nk}^T \mathbf{Y}_n$ is computed after
the QR decomposition.
The expression
\eqref{eq:M-kAR-theta} now can be computed by
\begin{math}
  \tilde{\Theta}_k^T =
  \left( \textstyle\sum_{n \in I_k} R_{nk}^T R_{nk} \right)^{-1}
  \left( \textstyle\sum_{n \in I_k} R_{nk}^T \mathbf{Y}_{Q_{nk}} \right).
\end{math}
The fast computation of \eqref{eq:M-theta} can be derived similarly.
Moreover, the VAR parameter for each time series $Y^{(n)}$ can be
computed by
\begin{math}
  \tilde{\Theta}_n^{*T} = R_{nk}^{-1} \mathbf{Y}_{Q_{nk}}.
\end{math}

Moreover, consider the QR decomposition 
\begin{math}
  \mathbf{E}_{nk} = U_{nk} V_{nk}
\end{math}
($n = 1,\dots,N; k = 1,\dots,K$),
where $U_{nk}$ is of dimension $(T-p) \times m$ and $V_{nk}$ of dimension
$m \times m$. Then $\Omega_k$ can be updated by
\begin{math}
  \Omega_k = \left( \textstyle\sum_{n \in I_k} V_{nk}^T V_{nk} \right)
  / \big( (T-p) |I_k| \big),
\end{math}
and the covariance estimation for each time series
$Y^{(n)}$, used for initialisation, can be similarly computed by
\begin{math}
  \Omega_n^* = \big( 1/(T-p_k) \big) V_{nk}^T V_{nk}.
\end{math}

To reliably compute \eqref{eq:E-kAR-psi}, consider a Cholesky
decomposition of $\Omega_k$,
\begin{math}
  \Omega_k = L_k L_k^T
\end{math}
($k = 1,\dots,K$),
where $L_k$ is a lower triangular matrix with real and positive diagonal
entries. The calculation of \eqref{eq:E-kAR-psi} can further rewritten as
\begin{math}
  \psi_{n,k} = \textstyle\sum_{t=p+1}^T (L_k^{-1} \mathbf{e}_{nkt})^T
  (L_k^{-1} \mathbf{e}_{nkt}),
\end{math}
where vector $L_k^{-1} \mathbf{e}_{nkt}$ is computed by solving
a system of linear equations $L_k x = \mathbf{e}_{nkt}$ w.r.t. $x$ due
to lower triangular $L_k$.


\subsection{Algorithm summary and parallelism}
\label{appdix:algorithm}

The complete pseudo-code of k-LMVAR algorithm is given by
Algorithm~\ref{alg:k-ARs}, with fast computation integrated. Several
computations in Algorithm~\ref{alg:k-ARs} can be parallised, including: (at
line 3) pre-computating $N$ number of QR decomposition, (at line 6-11)
computing $NK$ number of $\psi_{n,k}$, and (at line 12-16) updating $K$
number of parameters $\alpha_k, \tilde{\Theta}_k, \Omega_k$.

\begin{algorithm}[htbp]
  \caption{k-LMVAR algorithm}
  \label{alg:k-ARs}
  \begin{algorithmic}[1]
    \State \textsc{Input}: data $\mathbb{Y}$, number of clusters $K$,
    model orders $p_1,\dots,p_K$, and tolerance $\epsilon$.

    \State \textsc{Output}: model parameters $\tilde{\Theta}_k,
    \tilde{\Omega}_k$ (or $\Omega_k$), and clustering labels
    $\tau_{nk}$ ($k = 1,\dots,K$; $n=1,\dots,N$).

    \State Pre-computation: perform and cache QR decomposition of each
    $\mathbf{X}_{nk}$, and compute $\mathbf{Y}_{Q_{nk}}$.

    \State \textsc{Initialisation}: initialise $(\tilde{\Theta}_k, \Omega_k)$
    as Section~\ref{subsec:initialisation}.

    \While{TRUE}

    \For{$n \gets 1$ to $N$}
    \For{$k \gets 1$ to $K$}
    \State Compute $\psi_{n,k}$ by \eqref{eq:E-kAR-psi}.
    \EndFor
    \State Determine $\tau_{n,k}^{+}$ by \eqref{eq:E-kAR-tau} for $k=1,\dots,K$.
    \EndFor

    \For{$k \gets 1$ to $K$}
    \State Determine $I_k$ by $\tau_{n,k}^{+}$.
    \State Compute $\tilde{\Theta}_k^+$ by \eqref{eq:M-kAR-theta}.
    \State Compute $\Omega_k^+$ and $\tilde{\Omega}_k^+$ by  \eqref{eq:M-omega}.
    \EndFor

    \If{the stopping condition is satisfied}
    \State \textbf{break}
    \Else
    \State Update by
    $\tau_{n,k} \gets \tau_{n,k}^+, \tilde{\Theta}_k \gets
    \tilde{\Theta}_k^+, \Omega_k \gets \Omega_k^+$.
    \EndIf

    \EndWhile
  \end{algorithmic}
\end{algorithm}

\section{Model Selection}
\label{sec:model-selection}

With the benefits from the system identification approach, we are able to
use the BIC for the choice of the number of clusters and extra parameters.
To facilitate the selection of $K$ and $(p_1, \dots, p_K)$ for k-LMVAR, we
set up a general BIC criterion here. In practice it could useful to set
$p_1 = \cdots = p_K$. %
A direct application of BIC would fail due to that
the constant prior behind BIC amounts to assigning probabilities to model
classes associated with different $K$ proportional to their sizes. This is
strongly against the principle of parsimony, and may invalidate the Laplace
approximation in BIC.
The proper tool is the \emph{Extended BIC}, proposed in \cite{Chen2008}.

The whole model space $\mathcal{S}$ depends on both $K$ and
corresponding $(p_1, \dots,p_K)$. A natural partition of $\mathcal{S}$
is in accordance with the values of $K$. Assume $K$ can be any element
of an ordered set of $N_K$ integers, and each $p_k$ ($k=1,\dots,K$) can
any element of an ordered set of $N_p$ integers. Let the model set
$\mathcal{S}_j$ ($j = 1,\dots, N_K$) consist of any model with $K$ being
set to the $j$-th candidate of $K$, and $\mathrm{card}(\mathcal{S}_j)$
denote its size. Note that $p_k$'s may choose the same value, but, on
account of the nature of clustering, only their combination matters. For
instance, for $(p_1, p_2, p_3)$, $(1,1,2)$ and $(1,2,1)$ refer to the
same model. Thus, the size of model set $\mathcal{S}_j$ is the number of
$K$-combinations from a set of $N_p$ elements when repetition of
elements allowed,
\begin{math}
  \mathrm{card}(\mathcal{S}_j) = \binom{N_p+K-1}{K}
  = \frac{(N_p+K-1)!}{K!(N_p-1)!}.
\end{math}
The Extended BIC for the whole data $\mathbb{Y}$ is
\begin{equation*}
  \begin{array}{l@{\;}l}
    &\mathrm{BIC}_{\gamma}(K; p_1,\dots,p_K)  =
      -2 \log L + \Big[m^2 \big( \sum_{k=1}^K p_k \big) \\[4pt]
    &\quad + K \big( m^2/2 + 3m/2 \big) + \eta_{\mathrm{mix}}
        \Big] \log\big[ N(T-p_{\mathrm{max}}) \big] \\[4pt]
    &\quad + 2 \gamma \log \mathrm{card}(\mathcal{S}_j), \quad
      0 \leq \gamma \leq 1,
  \end{array}
\end{equation*}
where $\eta_{\mathrm{mix}} = K-1$ for cMVAR and
$\eta_{\mathrm{mix}} = N$ for k-LMVAR, $p_{\mathrm{max}}$ is
the maximal allowable order for $p_k$, and
\begin{equation*}
  \log L = \sum_{k=1}^K \sum_{n=1}^N \tau_{n,k}
  \Big[
  \frac{m}{2} (p_{\mathrm{max}}-T) \log(2\pi)
  -\frac{1}{2} D_{n,k}(\tilde{\Theta}_k, \Omega_k)
  \Big].
\end{equation*}
To avoid the varying length of
time series that is allowed for different models, we always use the time
points starting with $p_{\mathrm{max}}+1$ for predictions. As commented
in \cite{Chen2008}, the values of $\gamma=0, 0.5$ and $1$ are of special
interest. The value $0$ corresponds to the original BIC, where each
model is selected with an equal probability, while a model set
$\mathcal{S}_j$ with large $K$ has a high probability to be selected.
The value $1$ ensures each model set $\mathcal{S}_j$ is selected with an
equal probability and guarantee the consistency of the extended BIC. The
value $0.5$ ensures consistency when
$N_KN_p = O\big( (NT)^{\kappa} \big)$ with $\kappa < 1$, that is, the
value $N_KN_p$
is asymptotically upper bounded by $(NT)^{\kappa}$.

\section{Numerical Examples}
\label{sec:experiments}

In this section, we compare the proposed methods, {cMVAR} and {k-LMVAR},
with other state-of-the-art methods on synthetic datasets.


\subsection{Baseline Methods and Performance Indices}
\label{sec:basel-meth-perf}

The experiments will include the following state-of-the-art methods:
\begin{itemize}
\item k-DBA: k-Means with DTW barycenter averaging (DBA).
  We use \texttt{TimeSeriesKMeans} from the official Python package,
  \texttt{tslearn} \cite{tslearn}, with option \texttt{metric="dtw"}.

\item k-Shape: is proposed in \cite{Paparrizos2015,Paparrizos2017}, and we
  use \texttt{KShape} from the same package \texttt{tslearn}.

\item k-GAK: Kernel k-Means with Global Alignment Kernel \cite{Cuturi2011}
  based \cite{Dhillon2004} to perform time series clustering. We uses
  \texttt{KernelKMeans} from the package \texttt{tslearn}, with
  option \texttt{kernel="gak"}.

\item k-SC: k-Spectral Centroid proposed in \cite{Yang2011}. We use the
  MATLAB implementation given on the companion website of SNAP software
  \cite{Leskovec2014}.



\item DTCR: Deep Temporal Clustering Representation \cite{Ma2019a}. We use
  its official implementation in TensorFlow released on github.com.

\end{itemize}

Following the most recent literature \cite{Zhang2018b,Ma2019a}, the Rand
Index \cite{Rand1971} and Normalised Mutual Information \cite{Zhang2006}
are used to evaluate clustering performance:
\begin{align*}
  \mathrm{RI} &= \frac{\mathrm{TP} + \mathrm{TN}}{N(N-1)/2},\\
  \mathrm{NMI} &= \frac{\sum_{i=1}^K \sum_{j=1}^K N_{ij} \log \left(
      \frac{N N_{ij}}{|G_i| |A_j|} \right)}{\sqrt{
      \left( \sum_{i=1}^K |G_i| \log{\frac{|G_i|}{N}} \right)
      \left( \sum_{j=1}^K |A_j| \log{\frac{|A_j|}{N}} \right)
    }},
\end{align*}
where $\mathrm{TP}$ denotes True Positive, $\mathrm{TN}$ denotes True
Negative, $N$ is the total number of time series, $|G_i|$ and $|A_j|$
are the number of time series in cluster $G_i$ and $A_j$,
$N_{ij} = |G_i \cap A_j|$ denotes the number of time series belonging to
the intersection of sets $G_i$ and $A_j$, and $0 \log(0) = 0$ by
convention \cite{Cover2012}. Both of metrics with values close to 1
indicate high clustering performance.

\subsection{Experiments on clustering}
\label{subsec:synthetic-datasets}

To evaluate performance in a statistical manner, we randomly generate $40$
datasets for benchmark. Due to limitation of certain methods on
scalability, the problem size has to be restricted in the precision
benchmark such that all methods can function normally. Each time-series
dataset is generated by simulating $K$ number of VAR models using
\texttt{simulate} from MATLAB with the specifications as:
\begin{itemize}
  \vspace*{-1em}
\item the variable dimension: $m = 3, 6, 9$;
\item the order of the VAR model: $p = 5$;
\item the length of time series: $T = 100$;
\item the number of groups: $K = 8$;
\item the number of time series per cluster: $N_c = 40$, the total number
  of time series is $N = KN_c$.
\end{itemize}
\vspace*{-1em}
{\color{black} The random generation of each $\mathrm{VAR}(p)$ model
  requires stability being guaranteed, and $\Omega_k$ being positive
  definite. It is not a trivial task. Our approach is to first generate
  roots to construct stable characteristic polynomials and then calculate
  these parameters $\Theta_{k1},\dots,\Theta_{kp}$. And our experiment uses
  convariance matrix without any simplified structures as being diagonal or
  sparse.}

\begin{figure*}[htbp]
  \centering
  \includegraphics[width=.8\textwidth]{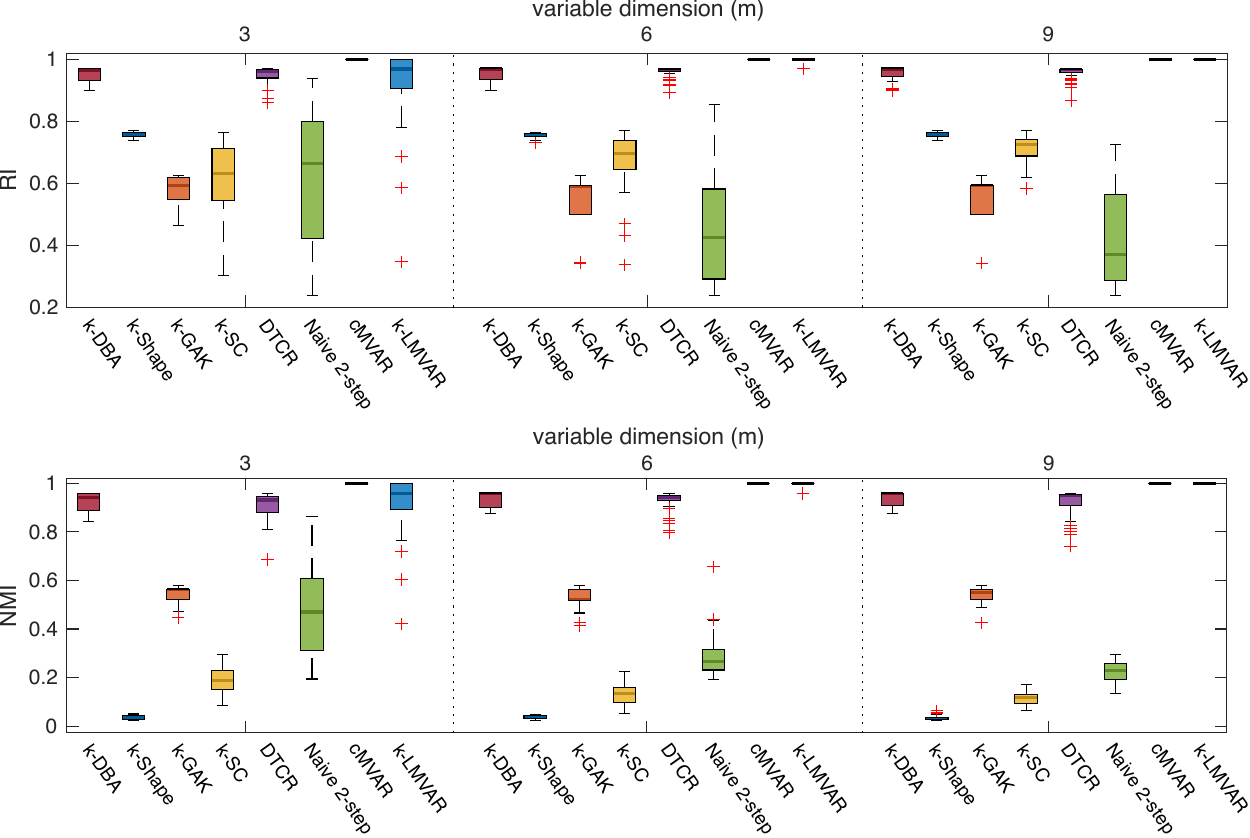}
  \caption{\color{black} Benchmark of clustering performance of cMVAR,
    k-LMVAR against the state-of-the-art, using Rand Index (RI) and
    Normalised Mutual Information (NMI).}
  \label{fig:bmk-clust-prec}
\end{figure*}

The benchmark of clustering precision, Figure~\ref{fig:bmk-clust-prec},
clearly show that the proposed methods, cMVAR and k-LMVAR, are superior
to the state-of-the-art methods on clustering precision. It may be
understood in such a way that our methods reply on dynamics or
statistical information of time series, rather than selected motifs
extracted by particularly designed discrepancy measures. It is obvious
that `shapelet' based methods fail to function well. The reason is that
the data are stochastic processes, whose characteristics are encoded in
the statistic. It may not be appropriate to assume that there always
exist consistently certain shape-like profiles that manage to
distinguish from each other. The DTCR method also shows satisfactory
performance, as a typical complicated black-box model, while it costs
a significant amount of time.

\subsection{Experiments on scalability}

To show the improvements of k-LMVAR on scalability and speed, we need
compare k-LMVAR with cMVAR in another setup of larger problem sizes.
The problem scales in two main ways:
\begin{itemize}
\item the number of time series: $N = N_c \times K$ ($N_c$ the
  number of time series per cluster), and
\item the dimension of variables $m$ or the time length $T$.
\end{itemize}
The former mainly shows the computational speed, where the scalability
may also be shown; and the latter shows the severe scalability issue
that cMVAR suffers.
To be specific, we have the following benchmark setups:
\begin{table}[H]
  \label{tbl:speed-setups}
  \centering
  \begin{tabular}{lrrrrr}
    \toprule
    & $m$ & $p$ & $T$ & $N_c$ & $K$ \\
    \midrule
    \textsc{Setup} 1 & 6 & 5 & 100 & 50 & 2:2:84 \\
    \textsc{Setup} 2a & 2 & 5 & 50:50:1200 & 20 & 5 \\
    \textsc{Setup} 2b & 2:1:20 & 5 & 150 & 20 & 5 \\
    \bottomrule
  \end{tabular}
\end{table}
where $a\!:\!i\!:\!b$ denotes a regularly-spaced vector from $a$ to $b$
using $i$ as the increment between elements. In both experiments, the
tolerance or numeric precision is set to $\epsilon = 10^{-8}$. And we
set the user-specified parameters $K$ and $p$ to their ground truth. To
generate the error plot (and balancing the total running time), we run 5
independent experiments for each method with every problem setup.

\begin{figure}[htb]   
  \begin{center}
    \centerline{\includegraphics[width=\columnwidth]{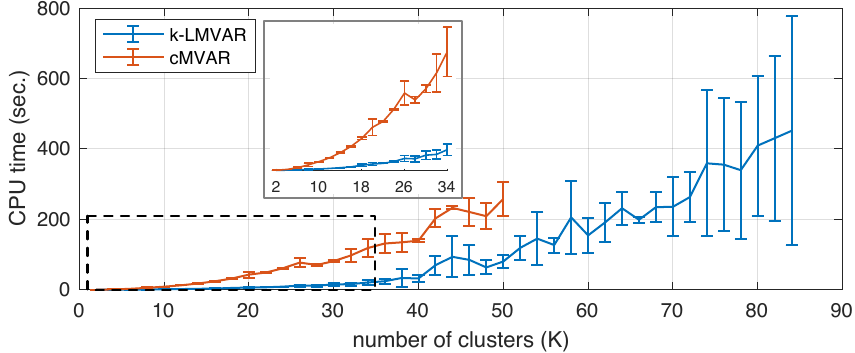}}
    \caption{Benchmark of computation time for k-LMVAR and cMVAR
      algorithms over different numbers of clusters. The curves
      boxed by dashed lines are zoomed in and shown in the floated
      sub-figure. The error bar illustrates one standard deviation.}
    \label{fig:bmk-clust-speed}
  \end{center}
\end{figure}

Figure~\ref{fig:bmk-clust-speed} mainly benchmarks the computation time
of k-LMVAR and cMVAR over increasing values of $K$ . It clearly shows
that k-LMVAR is consistently faster than cMVAR. What's more important,
with the increasing number of clusters, cMVAR fails to scale well and
shows the suffering from underflow issues. In this setup (i.e.
$m = 6, T = 100$), we have observed an increasing number of failures of
cMVAR due to underflow issues when $K$ goes above $40$. In a contrast,
k-LMVAR has no problem to handle the scale of much larger $K$ (here the
Monte Carlo experiment stops at $K=84$ due to unaffordable data size).
Our single test reveals that k-LMVAR can function well for $K$ as large
as $1000$ and more.

The increasing variance of computation time in
Figure~\ref{fig:bmk-clust-speed} when $K$ goes large can be understood
by learning the effects of initialisation. When $K$ is large, it implies
initialisation becomes more versatile. A poor initialisation means many
intermediate parameters pending to be updated, and thus yields a longer
iterative procedure to convergence.

\begin{figure}[htb]
  \centering

  \begin{subfigure}[b]{0.4\textwidth}
    \centering
    \includegraphics[width=\textwidth]{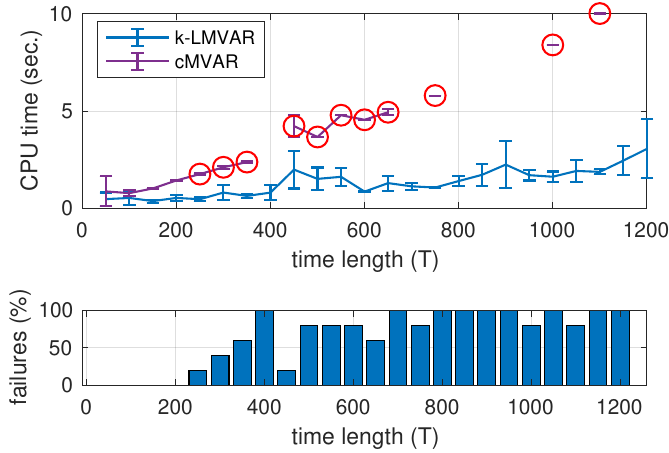}
    \caption{time length ($T$)}
    \label{subfig:scale-T-only}
  \end{subfigure}
  \\
  \begin{subfigure}[b]{0.4\textwidth}
    \centering
    \includegraphics[width=\textwidth]{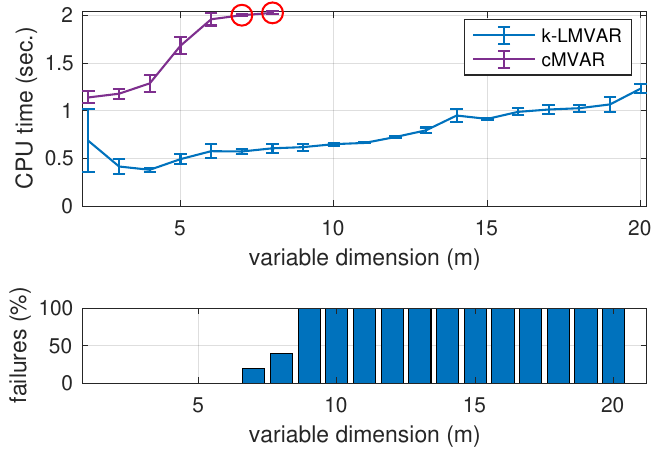}
    \caption{variable dimension ($m$)}
    \label{subfig:scale-m-only}
  \end{subfigure}

  \caption{Comparative study on scalability of k-LMVAR and cMVAR. The
    missing data points or points marked by red circles indicate the
    failures of cMVAR happening at the current problem setup. The missing
    of points of cMVAR in curves indicates the 100\% failure
    in the bar plots. The bar plots show the percentage of experiments
    that cMVAR fails due to its numerical issues.}
  \label{fig:scale-study-only}
\end{figure}

The comparative study on scalability of k-LMVAR and cMVAR is mainly shown
in Figure~\ref{fig:scale-study-only}. Here we present two scenarios: 1)
increasing $T$ (length of time series) with fixed $m=2$, and 2)
increasing $m$ (variable dimension) with fixed $T = 150$. In order to
get enough testing points for cMVAR, we have choose either a small $m$ or
$T$. Otherwise, it easily triggers the failure of cMVAR, for example,
even a small problem of $m=6, T=400$. Both
Figure~\ref{subfig:scale-T-only} and Figure~\ref{subfig:scale-m-only}
clearly show the weakness of cMVAR on scalability. In a contrast, k-LMVAR
can work without any issue on problems of at least 10 or 20 time larger
sizes in our test (note that we have not reached its performance limit).

\subsection{Examples of model selection}
\label{appdix:example-BIC}

\begin{figure}[htbp]   
  \centering
  \includegraphics[width=.48\textwidth]{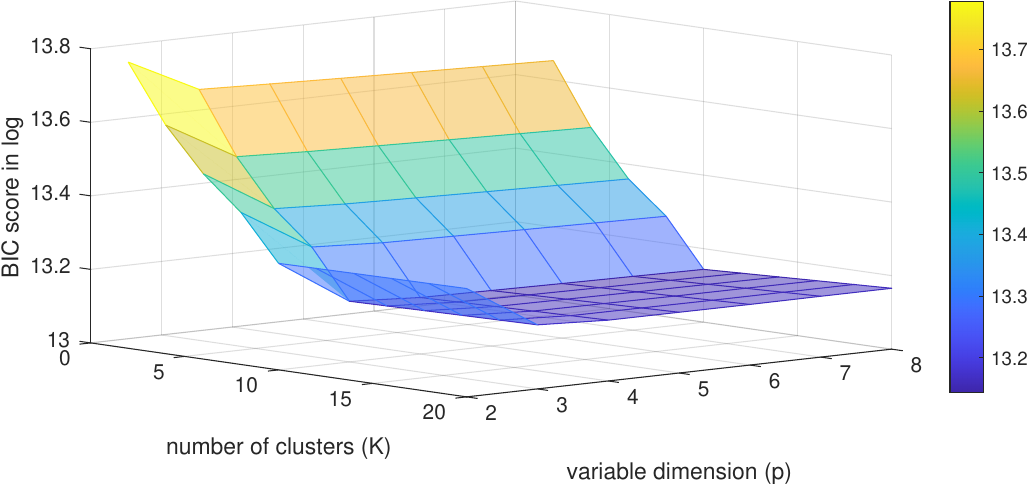}
  \caption{Mesh surface plot of BIC scores in terms of the number of
    clusters $K$ and model orders $p$.}
  \label{fig:BIC-mesh}
\end{figure}

An example of model selection by BIC is provided here. A set of time
series is generated from the set of models specified below:
\begin{itemize}
\item cluster-specific VAR models: $m = 4$, $p = 5$;
\item time series: $T = 200$, $K = 10$, $N_c = 20$ (number of time
  series per cluster).
\end{itemize}
The candidates for $K$ are $2\!:\!2\!:\!20$, and that for
$p$ are $2\!:\!1\!:\!8$. The values of $\mathrm{BIC}(K, p)$ over both $K$
and $p$ is visualised as a mesh surface plot in
Figure~\ref{fig:BIC-mesh}.

It is clear to see that the BIC score admits its minimal at the ground
truth $K=10$. The k-LMVAR algorithm is also a type of k-means-like
algorithm, which is not easy to see how BIC or AIC can be applied. Our
convergence analysis provides a construction of likelihood-like values
for applying BIC (note that it is not exactly the likelihood).

One may also observe that k-LMVAR algorithm is not sensitive to the
choice of model order $p$. In other words, a satisfactory clustering may
not depend on precise model order. This is different from modelling of
single time series. This implication is also supported by our benchmark
of clustering accuracy, where we are not required to select exact model
order to ensure performance. Model selection can also benefit from this
`property' of k-LMVAR: a coarse choice of model order has enabled us to
choose the number of clusters properly.

\section{Conclusion}
In this paper we have plugged a significant hole in the clustering
literature.
Most clustering algorithms fail to incorporate or omit the extraction of
the useful autocorrelation features. We began by applying an existing MVAR
model to develop an EM based clustering algorithm, cMVAR. However it is
limited in multivariate time series due to numerical underflow issues. By
taking a small noise limit we were able to derive a much simpler algorithm,
k-LMVAR, that can handle large variable dimension and time length. It
includes a BIC criterion for choosing tuning parameters. We illustrated the
new algorithm on simulated data showing its superiority on clustering
accuracy and scalability.

\begin{ack}
  This work was supported by the National Natural Science Foundation of
  China under Grant 52205520, and the Australian Research Council (ARC)
  Discovery Grants.
\end{ack}

\bibliographystyle{plain}
\bibliography{ref/library}

\appendix
\section{Supplementary Development Details}

\subsection{Algorithm development of cMVAR}
\label{appdix:deriv-EM}

\subsubsection{The complete-data log-likelihood}
\label{appdix:complete-data-log-lld}

Consider the latent variable $Z_n = (Z_{n,1},\dots,Z_{n,K})^T$ for
$n=1,\dots,N$, whose probability density is given by
\begin{equation*}
  p(Z_n) = \textstyle\prod_{k=1}^K \alpha_k^{Z_{n,k}},
\end{equation*}
where $\alpha_k$ is defined in the MVAR model. Note that $Z_{n,k}$ is
1-hot vector for each $n$. The conditional distribution of $Y^{(n)}_t$
given a particular value for $Z_n$ and its historical data is
\begin{equation*}
  p(Y^{(n)}_t \mid Z_{n,k} = 1,\; Y^{(n)}_{t-1}, \dots,
  Y^{(n)}_{t-p}) = f_k(\mathbf{e}_{nkt}, \Omega_k),
\end{equation*}
which can also be written in the form
\begin{equation*}
  p(Y^{(n)}_t \mid Z_{n}, Y^{(n)}_{t-1}, \dots, Y^{(n)}_{t-p})
= \textstyle\prod_{k=1}^K f_k(\mathbf{e}_{nkt}, \Omega_k)^{Z_{n,k}}.
\end{equation*}
Hence the `complete-data' distribution can be written as
\begin{equation*}
  \begin{array}{r@{\;}l}
    &p(Y^{(n)}_t, Z_{n} \mid Y^{(n)}_{t-1}, \dots, Y^{(n)}_{t-p}) \\[5pt]
    &=p(Y^{(n)}_t \mid Z_{n}, Y^{(n)}_{t-1}, \dots, Y^{(n)}_{t-p})
      p(Z_n) \\[5pt]
    &=\prod_{k=1}^K \left[ \alpha_k f_k(\mathbf{e}_{nkt}, \Omega_k)
      \right]^{Z_{n,k}}.
  \end{array}
\end{equation*}
By the product rule, it yields the `complete-data' likelihood of the
sequence $\{Y_t^{(n)}\}_{t=1}^T$ as follows
\begin{equation*}
  \begin{array}{r@{\;}r@{\;}l}
    \mathcal{L}^{(n)}
    &\coloneqq&
    p(Y^{(n)}_T, \dots,Y^{(n)}_{p+1}, Z_n \mid \tilde{\Theta}, \Omega) \\[5pt]
    &=&p(Y^{(n)}_T, Z_n \mid Y^{(n)}_{T-1},\dots,Y^{(n)}_{T-p}; \tilde{\Theta}, \Omega)
      \times \cdots \times \\[5pt]
    &&p(Y^{(n)}_{p+1}, Z_n \mid Y^{(n)}_{p},\dots,Y^{(n)}_{1};
      \tilde{\Theta}, \Omega) \\[5pt]
    &=&\prod_{t=p+1}^T \prod_{k=1}^K \left[ \alpha_k f_k(\mathbf{e}_{nkt}, \Omega_k)
      \right]^{Z_{n,k}}.
  \end{array}
\end{equation*}
Assuming that the $N$ number of time series are independently measured,
we then obtain \eqref{eq:lld-compl-data} via
\begin{equation*}
  \begin{array}{r@{\;}l}
    \log \mathcal{L}
    &= \log \left( \prod_{n=1}^N \mathcal{L}^{(n)} \right) \\[5pt]
    &= \log
    \left(
    \prod_{n=1}^N \prod_{t=p+1}^T \prod_{k=1}^K
    \left[
    \alpha_k f_k(\mathbf{e}_{nkt}, \Omega_k)
    \right]^{Z_{n,k}}
    \right)
  \end{array}
\end{equation*}
by altering the order of sums in terms of $n, k$ and $t$.

\subsubsection{Deviation of the EM steps}
\label{appdix:devitation-m-step}

The E-step evaluates the conditional probability of indicator variable
$Z_{n,k}$ given observation $Y^{(n)}$.
By Bayes' formula and the product rule of probability, we have
\begin{equation*}
  \begin{array}{r@{\;}l}
    \tau_{n,k} &= p\left( Z_{n,k} = 1 \mid Y^{(n)}; \tilde{\Theta},
                 \Omega \right) \\[5pt]
               &= \displaystyle\frac{p\left( Z_{n,k}=1 \right)
                 p\left( Y^{(n)} \mid Z_{n,k} = 1; \tilde{\Theta},
                 \Omega \right)}{\sum_{j=1}^K
                 p\left(Z_{n,j} = 1\right)
                 p\left( Y^{(n)} \mid Z_{n,j} = 1; \tilde{\Theta}, \Omega \right)}
  \end{array}
\end{equation*}
where
\begin{equation*}
  \begin{array}{r@{\;}r@{\;}l}
    &&p(Y^{(n)} \mid Z_{n,k} = 1; \tilde{\Theta}, \Omega) \\[5pt]
    &=& p(Y^{(n)}_T \mid Z_{n,k} = 1, Y^{(n)}_{T-1},\dots,Y^{(n)}_{T-p};
      \tilde{\Theta}, \Omega) \times \cdots \times \\[5pt]
    &&p(Y^{(n)}_{p+1} \mid Z_{n,k} = 1, Y^{(n)}_{p},\dots,Y^{(n)}_{1};
      \tilde{\Theta}, \Omega) \\[5pt]
    &=& \prod_{t = p+1}^T \alpha_k f_k(\mathbf{e}_{nkt}, \Omega_k).
  \end{array}
\end{equation*}
The expressions in the E-step \eqref{eq:E-tau} and \eqref{eq:E-psi} are
then obtained by reorganising the terms.

For the M-step, the key is to compute the derivatives of
$\log \mathcal{L}$ with respect to $\alpha_k$, $\tilde{\Theta}_k$ and
$\Omega_k$.
For $k = 1,\dots,K-1$, noting that
$\sum_{k=1}^K \alpha_k = 1$,
\begin{equation*}
  \frac{\partial \log \mathcal{L} }{\partial \alpha_k} =
  \sum_{n=1}^N \left( \frac{Z_{n,k}}{\alpha_k} - \frac{Z_{n,K}}{\alpha_K} \right).
\end{equation*}
For $k=1,\dots,K$,
\begin{align*}
  \frac{\partial \log \mathcal{L}}{\tilde{\Theta}_k}
  =
  & - \frac{1}{2} \Omega_k^{-1}
    \Big[
    \textstyle\sum_{n=1}^N Z_{n,k} \left( \sum_{t=p+1}^T \Theta_k X_{kt}^{(n)}
    X_{kt}^{(n)T} \right) \\[5pt]
  &-\textstyle\sum_{n=1}^N Z_{n,k} \left( \sum_{t=p+1}^T
    Y_t^{(n)} X_{kt}^{(n)T} \right)
    \Big], \\[5pt]
  \frac{\partial \log \mathcal{L}}{\tilde{\Theta}_k} =
  &\frac{1}{2}
    \Big[
    \Omega_k^{-1} \left( \textstyle\sum_{n=1}^N Z_{n,k} \sum_{t=1}^T
    \mathbf{e}_{nkt}\mathbf{e}_{nkt}^T \right) \\[5pt]
  &-(T-p) \textstyle\sum_{n=1}^N Z_{n,k}
    \Big] \Omega_{k}^{-1}.
\end{align*}
Setting the above derivatives to zero, and replacing $Z_{n,k}$ with
$\tau_{n,k}$, we immediately have \eqref{eq:M-alpha}-\eqref{eq:M-omega} for
the M-step of the cMVAR algorithm.

\subsection{Taking the Limit for k-LMVAR}
\label{appdix:calc-limit-vecKARs}

Recall that $\Omega_k = \gamma_k \tilde{\Omega}_k$ with
$\det(\tilde{\Omega}_k) = 1$, $\gamma = \max_k \gamma_k$, and all ratios
$\lambda_k = \gamma_k / \gamma$ are constants and $0 < \lambda_k \leq 1$.
For simplification, we introduce $M \triangleq -\frac{T-p}{2}$ and
$\beta_k \triangleq |\Omega_k|^{-M}$, and rewrite $\psi_{n,k}$
as follows
\begin{equation*}
  \begin{array}{r@{\;}l}
    \psi_{n,k}
    &= \gamma_k^{-1}
      \sum_{t=p+1}^T  \mathbf{e}_{nkt}^T \hat{\Omega}_k^{-1}
      \mathbf{e}_{nkt} \\[5pt]
    &= \gamma^{-1}
      \left( \lambda_k^{-1} \sum_{t=p+1}^T  \mathbf{e}_{nkt}^T
      \hat{\Omega}_k^{-1}  \mathbf{e}_{nkt} \right)
      \triangleq \gamma^{-1} \hat{\psi}_{n,k}.
  \end{array}
\end{equation*}
It is easy to see that $\hat{\psi}_{n,k}$ is bounded due to constant
$\lambda_k$ and bounded $\hat{\Omega}_k$. Let
$k^* \triangleq \argmin_{k'} \psi_{n,k'}$, and $\beta_{k^*}$,
$\psi_{n,k^*}$ and $\hat{\psi}_{n,k^*}$ be the corresponding values
when $k$ is set to $k^*$. Consider the limit
\begin{equation}
  \label{appdix:limit-tau}
  \begin{array}{r@{\;}l}
    \displaystyle\lim_{\gamma \rightarrow 0} \tau_{n,k}
    &= \displaystyle\lim_{\gamma \rightarrow 0}
      \frac{\alpha_k \beta_k \exp(-\frac{1}{2}\psi_{n,k})}{
      \sum_{k=1}^K \alpha_k \beta_k \exp(-\frac{1}{2}\psi_{n,k})} \\
    &= \displaystyle\lim_{\gamma \rightarrow 0}
      \frac{\alpha_k \frac{\beta_k}{\beta_{k^*}}
      \exp \left[ -\frac{1}{2 \gamma} \left( \hat{\psi}_{n,k} -
      \hat{\psi}_{n,k^*} \right) \right]}{
      \sum_{k=1}^K \alpha_k \frac{\beta_k}{\beta_{k^*}}
      \exp \left[ -\frac{1}{2 \gamma} \left( \hat{\psi}_{n,k} -
      \hat{\psi}_{n,k^*} \right) \right] },
  \end{array}
\end{equation}
where
$\frac{\beta_k}{\beta_{k^*}} =
\frac{|\hat{\Omega}_k|^{-M}}{|\hat{\Omega}_{k^*}|^{-M}} \left(
  \frac{\lambda_k}{\lambda_{k^*}}\right)^{-M} $ is bounded, and
$(\hat{\psi}_{n,k} - \hat{\psi}_{n,k^*})$ is bounded and non-negative since
it is equal to $\gamma (\psi_{n,k} - \psi_{n,k^*})$ that is non-negative.
Notice that the numerator of \eqref{appdix:limit-tau} approaches to zero
for $k \neq k^*$ when $\lambda \rightarrow 0$, but it is equal to a
non-zero constant for $k=k^*$. Now it is easy to see that the limit
\eqref{appdix:limit-tau} is equal to $1$ when $k=k^*$ and $0$ otherwise.

In practice, we do not have to compute the normalised
$\tilde{\Omega}_k$.
We may avoid the computation of the determinant, and use $\Omega_k$
directly in the E-step \eqref{eq:E-kAR-psi} without losing much
accuracy. Whereas, the computation of $\det(\Omega_k)$ may not be heavy
due to the available decomposition of $\Omega_k$ for the computation of
its inverse in \eqref{eq:E-psi}.

\section{Supplementary Convergence Analysis}
\label{appdix:lem-conv-analys}


We will provide basic convergence properties of the k-LMVAR algorithm.
By constructing a suitable optimization problem we are able to show
k-LMVAR is a coordinate (aka cyclic) descent algorithm
\cite{Wright2015,Luenberger2008}, which facilitates a discussion of
convergence.
Furthermore, the analysis of convergence is appreciated for implying a
construction of likelihood-like values to allow using information criteria.

Consider a mathematical program
\begin{equation}
  \label{eq:problem-P}
  \begin{array}{lc@{\;\;\;}l}
    P\text{:} & \underset{\tau, \tilde{\Theta},
                \tilde{\Omega}}{\mathrm{min}}\:
    & f(\tau, \tilde{\Theta}, \tilde{\Omega}) =
      \sum_{n=1}^N \sum_{k=1}^K \tau_{n,k} D_{n,k}(\tilde{\Theta}_k,
      \tilde{\Omega}_k) \\
              & \mathrm{s.t.}
    & \sum_{k=1}^K \tau_{n,k} = 1, \quad \tau_{n,k} = 0 \text{ or } 1, \\[1mm]
              & & \det(\tilde{\Omega}_k) = 1, \\[1mm]
    & & (k = 1,\dots,K; n = 1,\dots,N)
  \end{array}
\end{equation}
where $D_{n,k}(\tilde{\Theta}_k, \tilde{\Omega}_k)$ is an equivalent dissimilarity
measure constructed for k-LMVAR
\begin{equation}
  \label{eq:problem-P-dist-measure}
  D_{n,k}(\tilde{\Theta}_k, \tilde{\Omega}_k) = (T-p)\log |\tilde{\Omega}_k| +
  \textstyle\sum_{t=p+1}^T \mathbf{e}_{nkt}^T \tilde{\Omega}_{k}^{-1}  \mathbf{e}_{nkt}
\end{equation}
with $\mathbf{e}_{nkt}$ given in \eqref{eq:lld}; $\tau$ is a
$N \times K$ matrix of $\tau_{n,k}$\footnote{The variable $\tau_{n,k}$,
  in this section, corresponds to $\tau_{n,k}^*$ in \eqref{eq:M-tau-limit}
  and \eqref{eq:E-kAR-tau}. This section reserves the superscription
  `$*$' for optimal values.}; $\tilde{\Theta}$ denotes the set of all
$K$ parameters $\tilde{\Theta}_k$ given in \eqref{eq:lld}; and
$\tilde{\Omega}$ denotes the set of $\tilde{\Omega}_k$ ($k=1,\dots,K$) that are
all positive definite.

\begin{lem}
  \label{lem:partial-opt-sol}
  1) Given fixed $(\tilde{\Theta}^*, \tilde{\Omega}^*)$, the optimal
  solution of $\tau$ to $f(\tau, \tilde{\Theta}^*, \tilde{\Omega}^*)$ in
  \eqref{eq:problem-P} is given by \eqref{eq:E-kAR-tau}. 2) Given fixed
  $\tau^*$, the optimal solution of $(\tilde{\Theta}, \tilde{\Omega})$
  to $f(\tau^*, \tilde{\Theta}, \tilde{\Omega})$ in \eqref{eq:problem-P}
  is given by \eqref{eq:M-kAR-theta} and \eqref{eq:M-kAR-omega}.
\end{lem}

\begin{thm}
  \label{thm:convergence-f-seq}
  Let $\{\tau^{(l)}\}$, $\{\tilde{\Theta}^{(l)}\}$,
  $\{\tilde{\Omega}^{(l)}\}$ ($l = 1, 2, \dots$) be sequences generated
  by the k-LMVAR algorithm, and let
  $f^{(l)} \triangleq f(\tau^{(l)}, \tilde{\Theta}^{(l)},
  \tilde{\Omega}^{(l)})$ defined in \eqref{eq:problem-P}. Then the
  following results hold.
  \begin{enumerate}[label=\arabic*), leftmargin=2em]
  \item The sequence of cost function values $\{f^{(l)}\}$ ($l = 1,2,\dots$)
     is non-increasing and convergent.
  \item The k-LMVAR algorithm stops in a finite number of iterations.
  \end{enumerate}
\end{thm}

\subsection{Proof of Lemma~\ref{lem:partial-opt-sol}}
\label{appdix:lemma-part-optsol}

For simplicity, let $\theta \triangleq (\tilde{\Theta},
\tilde{\Omega})$, and hence the cost function $f(\tau, \tilde{\Theta},
\tilde{\Omega}) \triangleq f(\tau, \theta)$. The k-LMVAR algorithm
cycles between two simpler optimisations:
\begin{itemize}
\item Problem $P_{\tau}$: Given $\hat{\theta} \in S_{\theta}$, minimise
  $f(\tau, \hat{\theta})$ subject to $\tau \in S_{\tau}$;

\item Problem $P_{\theta}$: Given $\hat{\tau} \in S_{\tau}$, minimise
  $f(\hat{\tau}, \theta)$ subject to $\theta \in S_{\theta}$;
\end{itemize}
where the feasible sets are defined as follows
\begin{align*}
  S_{\tau}
  &\coloneqq \left\{
    \tau \in \{0,1\}^{N \times K} :
    \textstyle\sum_{k=1}^K \tau_{n,k} = 1
    \right\}, \\
  S_{\theta}
  &\coloneqq \left\{
    (\tilde{\Theta}, \tilde{\Omega}) :
    \det(\tilde{\Omega}_k) = 1, \;
    \tilde{\Omega}_k \text{ is positive definite},
    \forall k
    \right\}.
\end{align*}
Here we omit the trivial constraint that $\tilde{\Theta}$ is a
collection of real vectors/matrices of appropriate dimensions. To show
Lemma~\ref{lem:partial-opt-sol}, we simply need to show that 1)
\eqref{eq:E-kAR-tau}-\eqref{eq:E-kAR-psi} with given $\hat{\theta}$
solve Problem $P_{\tau}$; and 2)
\eqref{eq:M-kAR-theta}-\eqref{eq:M-kAR-omega} with given $\hat{\tau}$
solves Problem $P_{\theta}$.

The case of Problem $P_{\tau}$ is obvious by observing that any feasible
$\tilde{\Omega}$ (i.e. $\det(\tilde{\Omega}_k) = 1$) simplifies the
expression of the cost into
\begin{equation*}
  f(\tau, \hat{\theta}) =
  \textstyle\sum_{n=1}^N
  \left[
    \sum_{k=1}^K \tau_{n,k}
    \left(
      \sum_{t=p+1}^T\mathbf{e}_{nkt}^T \tilde{\Omega}_k^{-1} \mathbf{e}_{nkt}
    \right)
  \right].
\end{equation*}
It is now easy to see that the minimum of $f(\tau, \hat{\theta})$ is
achieved when, for each $n=1,\dots,N$, setting all $\tau_{n,k}$ to $0$
but the one with the minimal value of
$\sum_{t=p+1}^T\mathbf{e}_{nkt}^T \tilde{\Omega}_k^{-1}
\mathbf{e}_{nkt}$ (that is, $\psi_{n,k}$ in \eqref{eq:E-kAR-psi}).

For Problem $P_{\theta}$, since the cost function is separately in terms
of $k$, let us consider
\begin{equation*}
  \begin{array}{l@{\;\;}l}
    \underset{\theta_k}{\min}\:
    & f_k(\theta_k) \coloneqq \\
    & \sum_{n \in I_k}
      \left[
      (T-p) \log |\tilde{\Omega}_k| +
      \sum_{t=p+1}^T\mathbf{e}_{nkt}^T \tilde{\Omega}_k^{-1} \mathbf{e}_{nkt}
      \right] \\
    \mathrm{s.t.}
    & |\tilde{\Omega}_k| = 1.
  \end{array}
\end{equation*}
We use the method of Lagrange multiplier, which defines the Lagrangian
\begin{equation*}
  L(\theta_k, \lambda) = f_k(\theta_k) + \lambda \left( |\tilde{\Omega}_k| - 1 \right).
\end{equation*}
It is then to solve
$\nabla_{\tilde{\Theta}_k, \tilde{\Omega}_k, \lambda} L(\theta_k,
\lambda) = 0$. Solving $\partial L / \partial \tilde{\Theta}_k = 0$ yields
\eqref{eq:M-kAR-theta}. ${\partial L}/{\partial \lambda} = 0$ is
just the constraint $|\tilde{\Omega}_k| = 0$, and
\begin{equation*}
  \begin{array}{r@{\;}l}
    \displaystyle\frac{\partial L}{\partial {\tilde{\Omega}}_k} = 0
    \quad \Rightarrow \quad
    & |I_k| (T-p) \mathbf{I} -
    \sum_{n \in I_k} \sum_{t=p+1}^T \\
    &\tilde{\Omega}_k^{-T} \mathbf{e}_{nkt} \mathbf{e}_{nkt}^T +
    \lambda |\Omega_k| \mathbf{I} = 0,
  \end{array}
\end{equation*}
in which $\mathbf{I}$ is a matrix identity of the appropriate dimension,
and $|I_k|$ is the cardinality of set $I_k$.
Reorganising the above equation, it yields
\begin{equation*}
  \big( |I_k|(T-p) + \lambda |\tilde{\Omega}_k|  \big) \mathbf{I}
  = \tilde{\Omega}_k^{-1} \textstyle\sum_{n \in I_k} \sum_{t=p+1}^T
  \mathbf{e}_{nkt} \mathbf{e}_{nkt}^T,
\end{equation*}
assuming data sufficiency such that
$\sum_{n \in I_k} \sum_{t=p+1}^T \mathbf{e}_{nkt} \mathbf{e}_{nkt}^T$ is
invertible, whose solution can only be of the form
\begin{equation*}
  \tilde{\Omega}_k = \gamma \left( \textstyle\sum_{n \in I_k} \sum_{t=p+1}^T
    \mathbf{e}_{nkt} \mathbf{e}_{nkt}^T \right), \quad \text{with }
  \gamma \in \mathbb{R}^+.
\end{equation*}
Easily letting
\begin{math}
  \Omega_k = \sum_{n \in I_k} \sum_{t=p+1}^T \mathbf{e}_{nkt}
  \mathbf{e}_{nkt}^T
\end{math}
and $\gamma = \Omega_k / |\Omega_k|$ presents a solution. The expression
\eqref{eq:M-kAR-omega} keeps a scalar $1/[(T-p)|I_k|]$, which does not
affect the result but keeps a legacy relation to \eqref{eq:M-omega}.

\subsection{Proof of Theorem~\ref{thm:convergence-f-seq}}
\label{appdix:proof-theorem-2}

First we restate the k-LMVAR algorithm, in a form similar to a
k-means-like algorithm, as follows:
\begin{enumerate}[label=\arabic*)]
\item Choose an initial point $\theta^0$, solve ${P}_{\tau}$ with
  $\hat{\theta} = \theta^0$. Let $\tau^0$ be an optimal basic solution of
  ${P}_{\tau}$. Set $r = 0$.

\item Solve $P_{\theta}$ with $\hat{\tau} = \tau^r$. Let the solution be
  $\theta^{r+1}$. If $f(\hat{\tau}, \theta^{r+1}) = f(\hat{\tau},
  \theta^{r})$, stop and set $(\tau^*, \theta^*) = (\tau^*,
  \theta^{r+1})$; otherwise, go to step~3).

\item Solve $P_{\tau}$ with $\hat{\theta} = \theta^{r+1}$, and
  let the basic solution be $\tau^{r+1}$. If $f(\tau^{r+1},
  \hat{\theta}) = f(\tau^r, \hat{\theta})$, stop and set $(\tau^*,
  \theta^*) = (\tau^{r+1}, \hat{\theta})$; otherwise let $r= r+1$ and go
  to step~2).
\end{enumerate}

Let the generated sequence of cost functions $\{f^{(r)}\}_{r=1,2,\dots}$
be defined by
\begin{equation*}
  \begin{cases}
    f^{(2l)}   \triangleq f(\tau^l, \theta^l)
    & \text{when $r$ is even, } r = 2l, \\[4pt]
    f^{(2l+1)} \triangleq f(\tau^l, \theta^{l+1})
    & \text{when $r$ is odd, } r = 2l +1;
  \end{cases}
\end{equation*}
that is,
\begin{equation*}
  \begin{array}{r@{\;}l}
    &f^{(0)} \triangleq f(\tau^0, \theta^0), \quad
      f^{(1)} \triangleq f(\tau^0, \theta^1), \quad \\
    &f^{(2)} \triangleq f(\tau^1, \theta^1), \quad
      f^{(3)} \triangleq f(\tau^1, \theta^2), \quad \cdots.
  \end{array}
\end{equation*}
It is easy to see that the sequence $\{f^{(r)}\}$ is non-increasing from
the above description of k-LMVAR, without considering stopping
conditions. Moreover, since $\{f^{(r)}\}$ (without stopping applied) is
bounded lower by $0$, the sequence is convergent.

Now we can complete the  proof. First we
claim that a point in $S_{\tau}$ is visited at most once before
the algorithm stops. Suppose that this is not true, i.e.
$\tau^{r_1} = \tau^{r_2}$ for some $r_1 \neq r_2$. Considering the step
of `parameter update', we got optimal solutions $\theta^{r_1}$ and
$\theta^{r_2}$ for $\tau^{r_1}$ and $\tau^{r_2}$, respectively. It
immediately leads
\begin{equation*}
  f(\tau^{r_1}, \theta^{r_1}) = f(\tau^{r_2}, \theta^{r_1}) =
  f(\tau^{r_2}, \theta^{r_2}),
\end{equation*}
where the first equality is by $\tau^{r_1} = \tau^{r_2}$, and the second
is due to the step~2).
On the other hand, the sequence $f(\cdot, \cdot)$ generated by the
algorithm is strictly decreasing. A conflict rises and, hence the
assumption $\tau^{r_1} = \tau^{r_2}$ with $r_1 \neq r_2$ cannot be true.
Noticing that there are a finite number of elements in $S_{\tau}$, it
follows that the algorithm will stop in a finite number of iterations.




\section{Supplementary Model Selection}

\subsection{Basic case: the same order for all VARs}
\label{subsec:case-1-equal-orders}

Consider the special case where all $p_k$ ($k=1,\dots,K$) have the same
value $p$. When applying the clustering algorithms, it is necessary to
specify the two integer parameters $K$ and $p$. We can use the BIC criterion,
proposed in \cite{Schwarz1978}, to select $K$ and $p$ simultaneously.
BIC is then a surface, a function of two variables; computed as follows.
\begin{equation}
  \label{eq:BIC-K-p}
  \begin{array}{r@{\;}l}
    \mathrm{BIC}(K,p) =
    &- 2 \log {L} +
      \big\{ K \big[ (p+ 1/2) m^2 + 3m/2 \big]\\[5pt]
    &+\eta_{\mathrm{mix}}  \big\} \log\big[ N(T-p) \big] ,
  \end{array}
\end{equation}
where $\log{L}$ denotes the log-likelihood.
and $\eta_{\mathrm{mix}} = K-1$ for the MVAR-based algorithm and
$\eta_{\mathrm{mix}} = N$ for the k-LMVAR algorithm. For the MVAR-based
algorithm, the log-likelihood is simply given \eqref{eq:lld}.

However, for the k-LMVAR algorithm, we need a surrogate log-likelihood.
We get the following expression using the small noise heuristic.
\begin{equation*}
  \log L = \sum_{k=1}^K \sum_{n=1}^N \tau_{n,k}
  \Big[
  \frac{m}{2} (p-T) \log(2\pi)
  -\frac{1}{2} D_{n,k}(\tilde{\Theta}_k, \Omega_k)
  \Big],
\end{equation*}
with all $p_k$ in $D_{n,k}(\cdot, \cdot)$ being set to $p$ for
$\mathrm{MVAR}(m,K;p)$.

\subsection{Ad hoc Approach:  Separating selections of $K$ and VAR orders}
\label{subsec:separable-selection}

Alternatively, as suggested in \cite{Fong2007} for MVAR, we can
empirically choose $K$ and $p$ by BIC at two stages: first choose model
order $p$ via BIC for VAR modeling using one or a few time series, and
then run BIC with $p$ fixed to choose $K$. Indeed, this criterion
chooses $K$ and $p$ less `optimally', while it helps to avoid heavy
computation. Notice that, no matter which cluster a time series
$Y^{(n)}$ is assigned to, the order of VAR model estimation is the same.
The choice of model order $p$ is thought of as a task of VAR
modeling for one time series. Thus, we compute the BIC for VAR model
estimation using arbitrary time series $Y^{(n)}$, given by
\begin{equation}
  \label{eq:BIC-VAR}
  \mathrm{BIC}(p) = -2 \log L_{\mathrm{VAR}} +
  \Big[ m^2p + \frac{m(m+3)}{2} \Big] \log(T-p),
\end{equation}
where the likelihood of $Y^{(n)}$ is
\begin{displaymath}
  \log L_{\mathrm{VAR}} =
  -\frac{m}{2} (T-p) \log(2\pi)
  -\frac{1}{2} D_{n,k}(\tilde{\Theta}_k, \Omega_k)
\end{displaymath}
with index $k$ omitted. The model order $p$ is then chosen and fixed by the
value that minimises \eqref{eq:BIC-VAR}. For better reliability, one may
repeat multiple times using different time series in $\mathbb{Y}$ and
then select the value of $p$ that appears the most times. Once model order
$p$ is fixed, we can use \eqref{eq:BIC-K-p} to calculate the BIC for
clustering algorithms to determine the number of clusters $K$.


The above suggestion from \cite{Fong2007} is just a special case of such
an \emph{ad hoc} approach that optimises BIC using cyclic descent. Given
$K$ minimize BIC with respect to $p$; Given $p$ minimize BIC with
respect to $K$. Then cycle between these two steps.
The case of $p_k = p, \forall k$ is simple, and one may only need to
analyse a few time series to choose the value of $p$. Whereas, for the
general case of $p_k, k=1,\dots,K$, one may have to study a large enough
number of time series in order to exhaust all possible values of
$p_k$'s. It is still beneficial to computation costs, in comparison to
the heavy computation of application of extended BIC in grid search.


\section{Random Generation and Simulation of Stable VAR Models}
\label{appdix:rand-gener-stable-VAR}

To compare clustering methods we simulate a set of stable VAR models to
construct synthetic datasets. The basic specifications are given in
Section~\ref{subsec:synthetic-datasets}, including $m, p, T$ and $K, N$.
Here we will present more details on model construction and simulation.

Consider a model $\mathrm{MVAR}(m, K; p_1,\dots,p_K)$, which consists of
$K$ number of VAR models $\mathrm{VAR}(p_k)$ ($k=1,\dots,K$). Each
model $\mathrm{VAR}(p_k)$ has the following parameters pending to be
constructed:
\begin{math}
  \Theta_{k0}, \Theta_{k1}, \dots, \Theta_{kp_k}, \Omega_k,
\end{math}
where, in addition, the set of $\{\Theta_{k0}, \dots, \Theta_{kp_k}\}$
is required to guarantee its VAR being stable, and $\Omega_k$ is
required to be positive definite. For clearance of notations, we will
focus on one VAR model $\mathrm{VAR}(p)$ and omit the index $k$
without ambiguity. It consists of two tasks:
\begin{itemize}
\item randomly generating $\Theta_0, \Theta_1, \dots, \Theta_p$ such
  that $\mathrm{VAR}(p)$ is stable; and

\item randomly generating symmetric positive definite $\Omega$.
\end{itemize}

Recall the stability condition for $\mathrm{VAR}(p)$ as follow (e.g., see
\cite{Lutkepohl2005}):
\begin{equation}
  \label{eq:VAR-stab-cond}
  \det\left( I - \Theta_1 z - \cdots - \Theta_p z^p \right) \neq 0
  \quad \text{for any } |z| \leq 1, z \in \mathbb{C},
\end{equation}
i.e. the reverse characteristic polynomials has no roots in and on the
complex unit circle. The onus is to find a way to construct AR
parameters that sufficiently guarantee \eqref{eq:VAR-stab-cond}. Let us
assume the $\Theta_i$'s ($i=1,\dots,p$) admit a simultaneous
eigendecomposition, that is,
\begin{equation}
  \label{eq:simul-eigen-decomp}
  \Theta_i = U^{-1} \Lambda_i U \quad \text{ for all } i = 1,\dots,p,
\end{equation}
where $\Lambda_i = \mathrm{diag}(\lambda_i^{(1)}, \dots, \lambda_i^{(m)})$.
Then, the condition \eqref{eq:VAR-stab-cond} for such a family of
$\Theta_i$'s can be equivalently rewritten as
\begin{equation}
  \label{eq:VAR-stab-cond-eig}
  g_j(z) \triangleq 1 - \lambda_1^{(j)} z - \cdots - \lambda_p^{(j)} z^p \neq 0
  \quad \text{for any } |z| \leq 1
\end{equation}
holds for $j = 1,\dots,m$. Now it is easy to construct AR parameters to
satisfy \eqref{eq:VAR-stab-cond-eig}, as follows:
\begin{enumerate}[label=\arabic*)]
\item Randomly generating $mp$ number of real numbers with absolute
  value larger than $1$, denoted by $\{z_1^{(j)}, \dots, z_p^{(j)}\}$
  ($j=1,\dots,m$), which will used as the roots of $g_j(z) = 0$.

\item Construct the polynomial $g_j(z)$ by roots $\{z_1^{(j)}, \dots,
  z_p^{(j)}\}$, and then obtain the coefficients $\{\lambda_1^{(j)},
  \dots, \lambda_p^{(j)}\}$, for each $j=1,\dots,m$.

\item Randomly generate a $m \times m$ matrix $X$, and perform QR
  decomposition on $X$ to obtain its corresponding orthogonal matrix
  $U$, which will be used for \eqref{eq:simul-eigen-decomp}.

\item For each set $\{\lambda_i^{(1)}, \dots, \lambda_i^{(m)}\}$,
  construct $\Lambda_i$ and then we have $\Theta_i$ ($i = 1,\dots,p$) by
  \eqref{eq:simul-eigen-decomp} such that the stability condition
  \eqref{eq:VAR-stab-cond} is guaranteed.
\end{enumerate}
In regard to $\Theta_0$, it does not affect the stability and thus can
be randomly generated.

To generate $\Omega$, we use the property that a symmetric matrix is
positive definite if and only it can be factored into a product $L^TL$
for some nonsingular matrix $L$. Thus, we simply generate a nonsingular
matrix $L$ randomly and then set $\Omega = L^T L$.

To general time series from VAR models, our implementation uses the
standard function \texttt{simulate} from \emph{Econometric Toolbox} of
MATLAB, which does Monte Carlo simulation of VAR models.

\section{Additional Experimental Results}
\label{sec:add-experiments}

\begin{figure}[htbp]   
  \centering
  \includegraphics[width=\columnwidth]{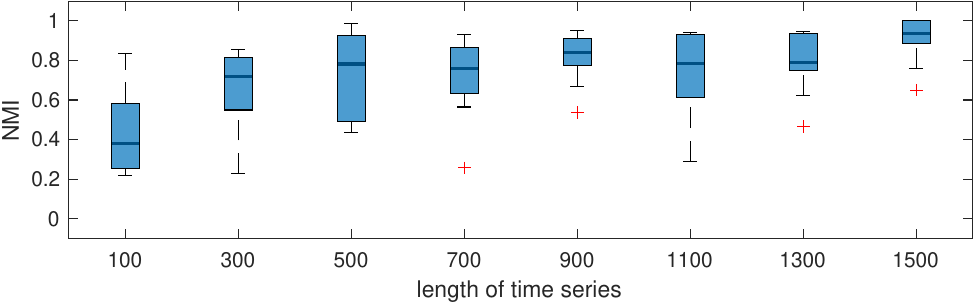}
  \caption{Clustering performance (using NMI) of the naive two-step
    method that applies k-means on the VAR model parameters estimated
    for each time series, given the different length of time series.}
  \label{fig:perf-two-step-method}
\end{figure}

As clearly seen in Figure~\ref{fig:bmk-clust-prec}, the performance of
the naive two-step approach is quite poor as we predicated in
Section~\ref{sec:prob-descrip}. However, this naive idea should at least
work well under `ideal' setups, which is explored in
Figure~\ref{fig:perf-two-step-method}. This supplementary experiment
follows the previous setup except allowing up to 15 times longer time
length, where $T$ can be 100 to 1500 with an increment of 200. As
illustrated in Figure~\ref{fig:perf-two-step-method}, when the length of
time series gets larger, its performance on average is increasing. This
is easy to foreseen since the estimation of VAR parameters is getting
better. On the other side, this experiment reveals another advantage of
our methods (cMVAR and k-LMVAR): the clustering can be done well even if
modelling of each time series is poor due to deficit data. The reason is
that such methods actually benefit from the whole data in essential
estimation of cluster-specific models. Thus, we are applying the system
identification approach to the time-series clustering problem, while the
application needs to be smart for sound performance.

\end{document}